\documentclass{svproc}
\usepackage{url}

\usepackage{cite}
\usepackage{stmaryrd}
\usepackage{amsmath,amssymb,amsfonts}
\usepackage{algorithmic}
\usepackage{graphicx}
\usepackage{textcomp}
\usepackage{float}
\usepackage{bm}
\usepackage{url}            
\usepackage{booktabs}       
\usepackage{amsfonts}       
\usepackage{nicefrac}       
\usepackage{microtype}      
\usepackage{lipsum}		
\usepackage{graphicx}
\usepackage{multicol}
\usepackage{algorithm}
\usepackage{subcaption}
\usepackage[square,sort,comma,numbers]{natbib}

\usepackage{braket}

\usepackage{bera}
\usepackage{listings}
\usepackage{xcolor}

\colorlet{punct}{red!60!black}
\definecolor{background}{HTML}{EEEEEE}
\definecolor{delim}{RGB}{20,105,176}
\colorlet{numb}{magenta!60!black}
\begin{document}

\lstdefinelanguage{json}{
    basicstyle=\footnotesize\ttfamily,
    numbers=left,
    numberstyle=\scriptsize,
    stepnumber=1,
    numbersep=6pt,
    showstringspaces=false,
    breaklines=true,
    frame=lines,
    backgroundcolor=\color{background},
    literate=
     *{0}{{{\color{numb}0}}}{1}
      {1}{{{\color{numb}1}}}{1}
      {2}{{{\color{numb}2}}}{1}
      {3}{{{\color{numb}3}}}{1}
      {4}{{{\color{numb}4}}}{1}
      {5}{{{\color{numb}5}}}{1}
      {6}{{{\color{numb}6}}}{1}
      {7}{{{\color{numb}7}}}{1}
      {8}{{{\color{numb}8}}}{1}
      {9}{{{\color{numb}9}}}{1}
      {:}{{{\color{punct}{:}}}}{1}
      {,}{{{\color{punct}{,}}}}{1}
      {\{}{{{\color{delim}{\{}}}}{1}
      {\}}{{{\color{delim}{\}}}}}{1}
      {[}{{{\color{delim}{[}}}}{1}
      {]}{{{\color{delim}{]}}}}{1},
}
\mainmatter              

\title{A derivation of variational message passing (VMP) for latent Dirichlet allocation (LDA)}
\titlerunning{A derivation of VMP for LDA}  
%
%

%
\author{Rebecca M.C.~Taylor\inst{1,2} \and Dirko Coetsee\inst{1,2,3} \and Johan A. du Preez\inst{1}}
%
%
%
\institute{Stellenbosch University, South Africa
\and 
Praelexis, South Africa
\and KU Leuven, Belgium}

\maketitle              

\begin{abstract}
Latent Dirichlet Allocation (LDA) is a probabilistic model used to uncover 
latent topics in a corpus of documents. 
Inference is often performed using variational Bayes (VB) algorithms,
which calculate a lower bound to the posterior distribution over the 
parameters.
Deriving the variational update equations for new models requires considerable manual effort; variational message passing (VMP) has emerged as a
"black-box" tool to expedite the process of variational inference.
But applying VMP in practice still presents subtle challenges, 
and the existing literature does not contain the steps that are necessary 
to implement VMP for the standard smoothed LDA model,  nor are available black-box probabilistic graphical modelling software able to do the word-topic updates necessary to implement LDA.
In this paper, we therefore present a detailed derivation of the VMP update 
equations for LDA. We see this as a first step to enabling other researchers to calculate the VMP updates for similar graphical models.

\end{abstract}
\section{Introduction}
\label{sec:Introduction}
Latent Dirichlet Allocation (LDA) \cite{blei2003latent} is an effective 
and popular probabilistic document model with many applications 
\cite{jelodar2019latent}. These include,
\begin{description}
  \item[\hspace{0.5cm}banking and finance:] clustering banking clients based on their 
   transactions and identification of insurance fraud \cite{liu2012identifying}.
  \item[\hspace{0.5cm}genomics:] using multilocus genotype data to learn about population 
   structure and assign individuals to populations \cite{pritchard2000inference},
   classification of gene expression in healthy and diseased tissues,
   \cite{yalamanchili2017latent} and prediction of the functional effects of 
   genetic variation \cite{backenroth2018fun},
  \item[\hspace{0.5cm}image processing:] image clustering and scene recognition 
   \cite{fei2005bayesian,cao2007spatially}, and
  \item[\hspace{0.5cm}medical:] feature extraction for rare and emerging diseases 
   \cite{gupta2021pan} and medical data set clustering 
   \cite{joshi2020modified,selvi2019classification}.
\end{description}
While LDA can extract latent topics of any type from a wide range of inputs,
it is most commonly used to extract latent semantic information from text.  
The scale at which LDA is applied has continued to grow with the increasing 
availability of computing resources, large volumes of data
\cite{masegosa2017scaling}, and improved inference algorithms.

\begin{figure}[h]
  \vskip 0.2in
  \begin{center}
    \centerline{\includegraphics[width=0.8\columnwidth]{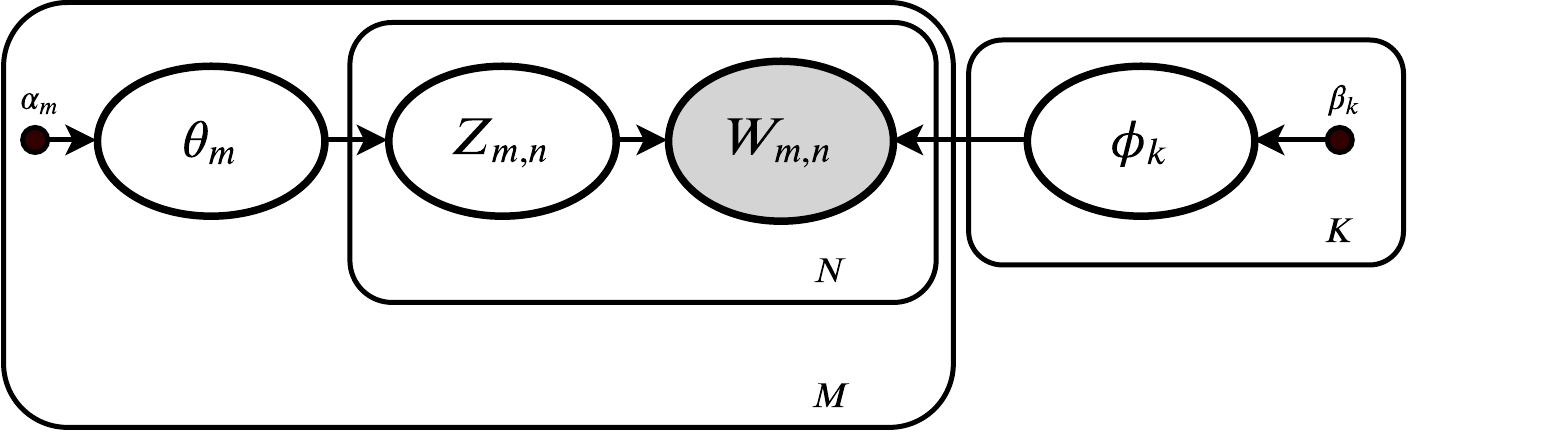}}
    \caption{Plate model for LDA as a Bayesian network. 
     Each node in the graph represents a random variable and edges represent 
     conditional dependencies.
     \label{fig:bn}}
  \end{center}
  \vskip -0.2in
\end{figure}

LDA is usually represented as a graphical model, as shown in Figure \ref{fig:bn}.
This graph represents a probability distribution, where $M$ documents 
contain $N$ words, and each word represents one of $K$ possible topics.
Section \ref{sec:ldagraph} provides further details.
LDA has been extended and modified to create many new but similar 
graphical models such as Filtered-LDA \cite{alattar2021emerging},
author topic models \cite{steyvers2004probabilistic},
relational topic models \cite{chang2009relational},
dynamic topic models \cite{blei2006dynamic}, and 
spatial LDA \cite{wang2007spatial}.

Exact inference is computationally intractable for many useful graphical models, 
including LDA and its many variants 
\cite{blei2003latent, NIPS2001_296472c9, blei2017variational}, 
\cite[p461]{bishop2006pattern}. 
A range of approximate inference techniques is therefore used.
Particle based approaches such as Markov chain Monte Carlo (MCMC) 
\cite[p462]{bishop2006pattern} have been used but are still 
computationally expensive \cite{wainwright2008graphical}.

Because larger data sources are now readily available,
faster and comparably accurate methods to these particle-based approaches have 
gained popularity \cite{blei2017variational}.
These include variational Bayes (VB) 
\cite{attias2000variational,asuncion2009smoothing,braun2010variational} and 
expectation propagation (EP) 
\cite{minka2001expectation,minka2001family,minka2004power}.
Variational Bayes in particular is notable for achieving good run-time 
performance. By optimising a bound on the posterior distribution, it provides guarantees on the inference quality that is not always attainable with 
approximate methods.
VB is usually available only for exponential family distributions. In Section 3.2 we will therefore provide a short overview of the exponential family and some of the notation we 
use later on in this paper.

Constructing a new model that uses VB is labour-intensive since the modeller must derive each of the variational update equations manually.  This process has been somewhat eased by the introduction of 
variational message passing (VMP), a formulation of VB that promises to be 
general enough to apply to a wide variety of graphical models but also a 
"black-box" inference engine that automates the calculation of the update 
equations \cite{blei2017variational}. A general overview of VMP is provided in Section~\ref{sec:vmp}.

The available VMP toolkits, however, do not cater for all possible
conditional probability distributions, either because a 
specific distribution is not implemented yet or would be too costly in terms of computational or memory resources.

The conditional Dirichlet-multinomial distribution in particular is necessary 
to implement smoothed LDA, as discussed in Section \ref{sec:vpmlda} but,
to the best of our knowledge, is not found in any of the popular toolkits~\cite{tensorflow2015-whitepaper,InferNET18}.
As a first step towards incorporating this distribution, and as an aid to other 
researchers using similar models, we 
derive the full VMP update equations for LDA in Section \ref{sec:vpmlda}.
This is the first time, to the best of our knowledge, that these equations have been published.

\section{Related work}
\label{sec:vb}
A goal of probabilistic modelling is to differentiate between the model and
the inference algorithm to answer queries about the model.
Concerning the modelling aspect, two prominent LDA models exist, a non-smoothed and a smoothed 
version~\cite{blei2003latent}. The smoothed version, which is considered here, has become the standard
version for topic modelling, and we will only consider it further here.

Most related to our work is the inference aspect of probabilistic modelling, particularly the different inference techniques that have been 
proposed for LDA.
 As far as inference is concerned, there is generally a trade-off between inference quality, speed of execution, 
and guarantees such as whether converge is guaranteed.
Below we mention some of the inference techniques and how they relate to VMP.

Collapsed Gibbs sampling (a type of MCMC technique) is often the 
inference technique of choice for LDA because it is theoretically exact in the 
limit. Although it provides high quality inference results,
it often requires a prolonged run-time for LDA.
 
Variational Bayesian inference is not exact, but is usually faster, 
is guaranteed to converge, and provides some guarantees on the quality of the 
result.
The smoothed version of LDA was 
first introduced using variational Bayesian inference \cite{blei2003latent}.
An online version was later introduced in 2011 \cite{wang2011online},
and a stochastic variant in 2013 \cite{hoffman2013stochastic}.
Both of these later methods are beneficial when there are larger amounts of data,
but they do not improve LDA performance as much as other later methods 
\cite{wang2011online,hoffman2013stochastic,asuncion2009smoothing,sato2012rethinking}.

Structured stochastic variational inference, 
introduced in 2015 \cite{hoffman2015structured},
further improved performance and scalability. 
Standard VB, however, is still a popular technique for LDA due to its simplicity 
and many Python implementations.

Variational message passing (VMP) 
\cite{winn2004variational, winn2005variational} is the message passing 
equivalent of the standard version of VB. 
It is a useful tool for constructing a variational inference solution for a 
large variety of conjugate exponential graphical models.
A non-conjugate variant of VMP, namely non-conjugate VPM (NCVMP)
is also available for certain other models \cite{knowles2011non}, 
but not applicable to this work.
An advantage of VMP is that it can speed up the process of deriving a 
variational solution to a new graphical model \cite{blei2017variational}.

There are software toolkits that implement VMP for general graphical models \cite{tensorflow2015-whitepaper,InferNET18}, but none, as far as we can tell, are able to do VMP 
for LDA-type models at the moment. 
For Infer.NET, arguably the most well-known VMP toolkit, the reason is that the 
current software implementation 
stores all intermediate messages as separate objects in memory, and the VMP 
messages for a typical Dirichlet-multinomial distribution used in LDA would take 
too much memory. 
Although there are no immediate plans to change the implementation to allow 
these updates~\cite{commMinka}, we believe that doing this could be fruitful future work.

Unfortunately, VMP update equations are therefore sometimes derived by hand,
but these results are usually not published \cite{masegosa2016d, masegosa2017scaling}, in contrast to 
the current work.

\section{Background}
In this section we present the graphical model for LDA, and also introduce the exponential family.
\subsection{The latent Dirichlet allocation (LDA) graphical model}
\label{sec:ldagraph}

Latent Dirichlet Allocation (LDA) is a hierarchical graphical model 
that can be represented by the directed graphical model 
shown in Figure \ref{fig:bn} \cite{blei2003latent} (see Table~\ref{table:symbols} for details about the symbols).

The graphical model shown in Figure \ref{fig:bn} allows us to visually identify 
the conditional independence assumptions in the LDA model. 
Arrows in the graph indicate the direction of dependence.
From Figure~\ref{fig:bn}, we can see that the $n$'th word in document $m$ is $W_{m,n}$.
The distribution over this word depends on the topic $Z_{m,n}$ present in the 
document, which, selects the Dirichlet random vector $\mathbf{\phi}_{k}$ that 
describes the words present in each topic.

\begin{figure}[h]
  \includegraphics[width=0.5\linewidth]{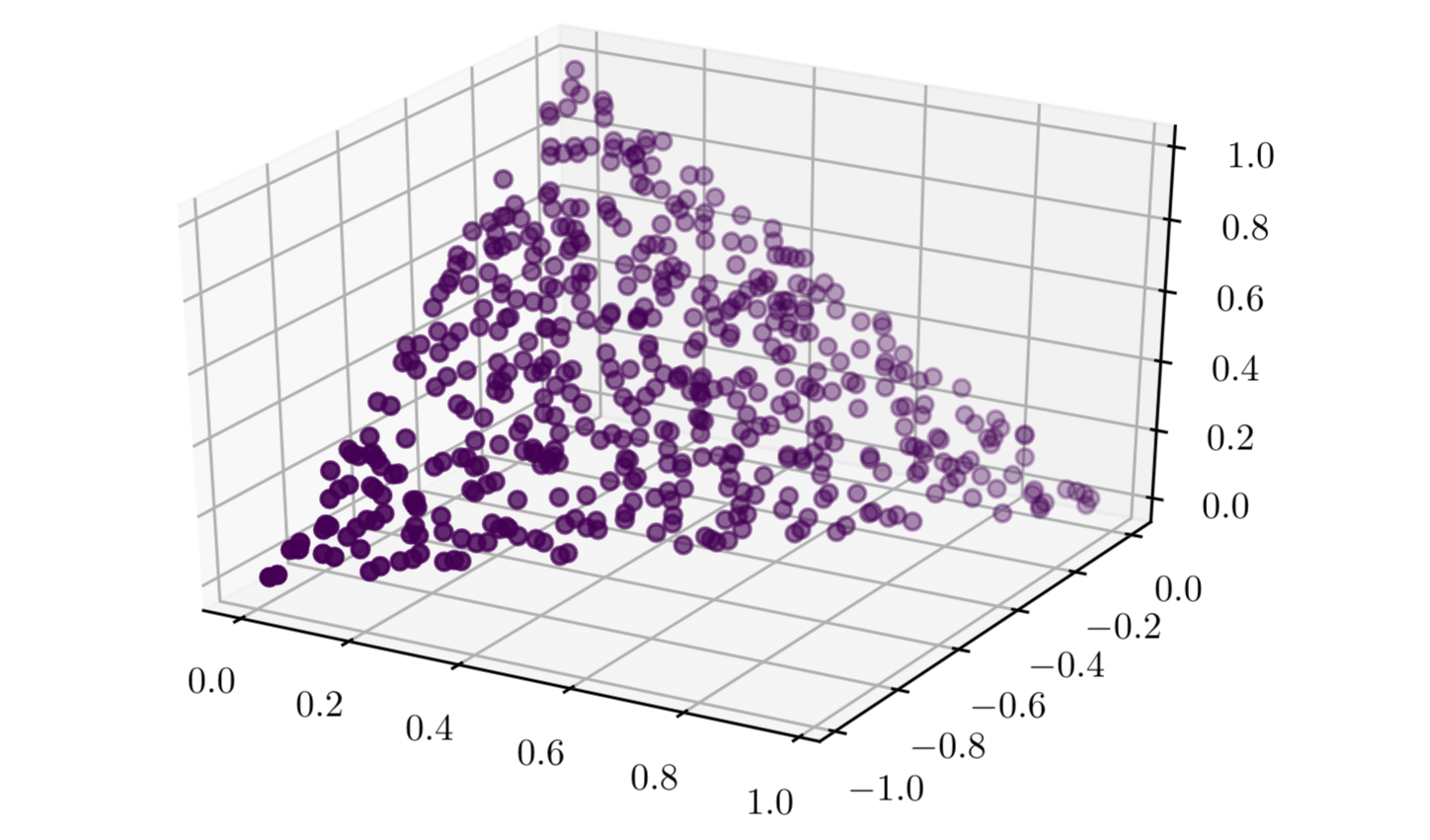}
  \includegraphics[width=0.5\linewidth]{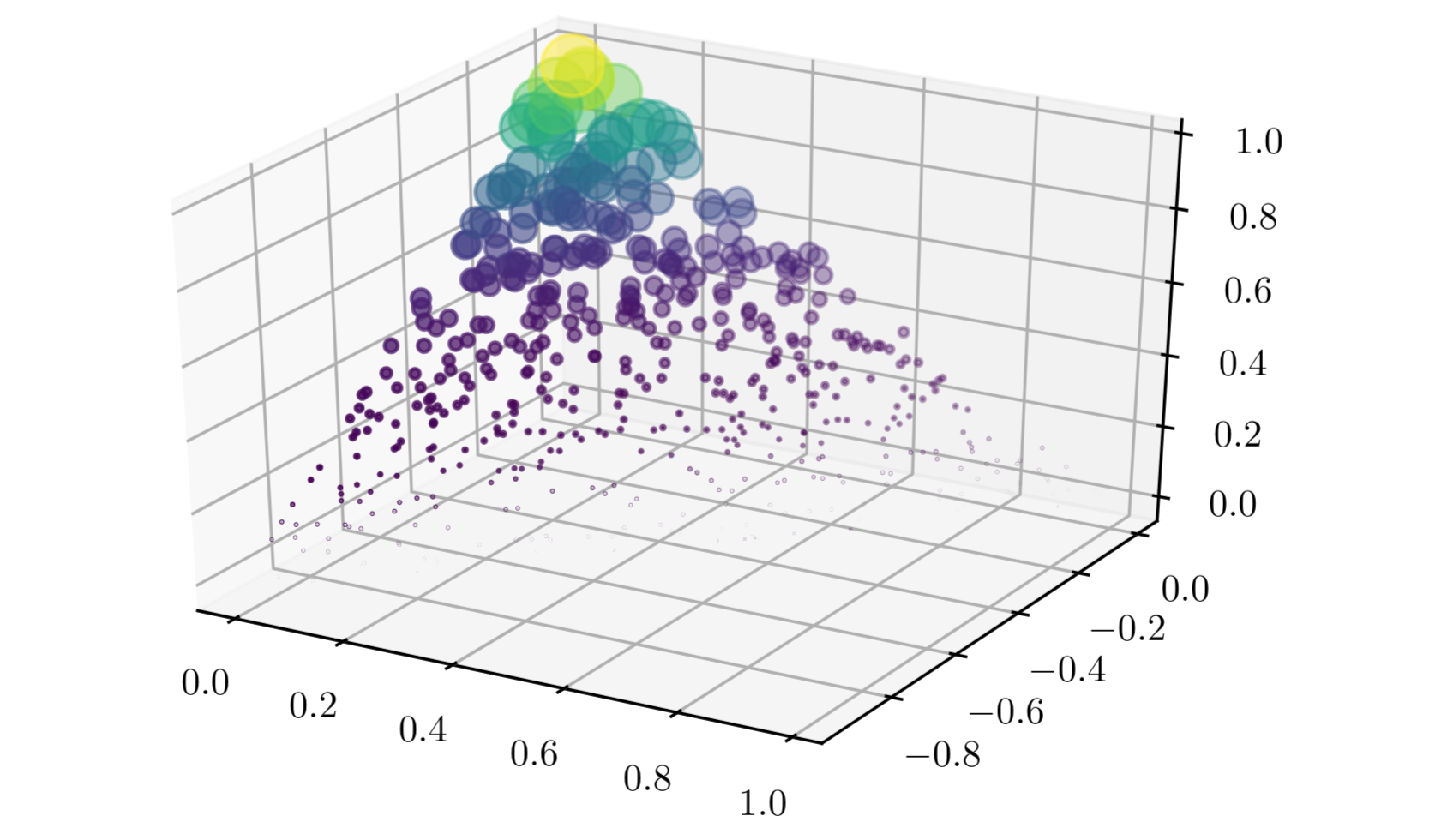}
  \caption{
    Dirichlet distributions of cardinality, $3$, visualised in 3D.
    (\textbf{a}) Dirichlet with $\bm{\alpha} = \{1,1,1\}$. 
    This is the non-informative Dirichlet.
    (\textbf{b}) Dirichlet with $\bm{\alpha} = \{1,1,5\}$.
    There is more mass on the corner of the Z-axis due to the higher 
    pseudocount of $\alpha_3$.
    \label{fig:dirsa}}
\end{figure}

\begin{figure}[h]
  \includegraphics[width=0.5\linewidth]{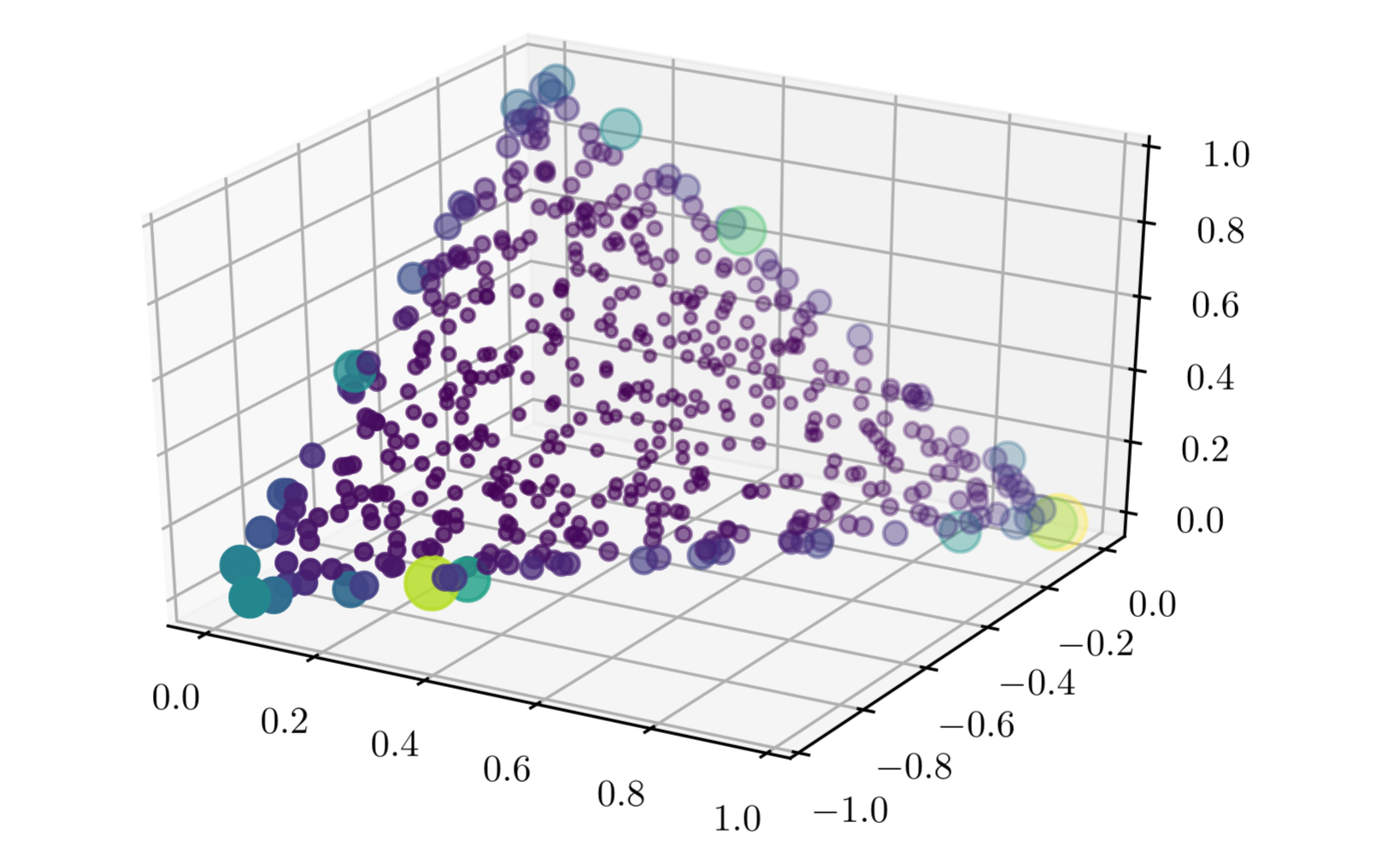}
  \includegraphics[width=0.5\linewidth]{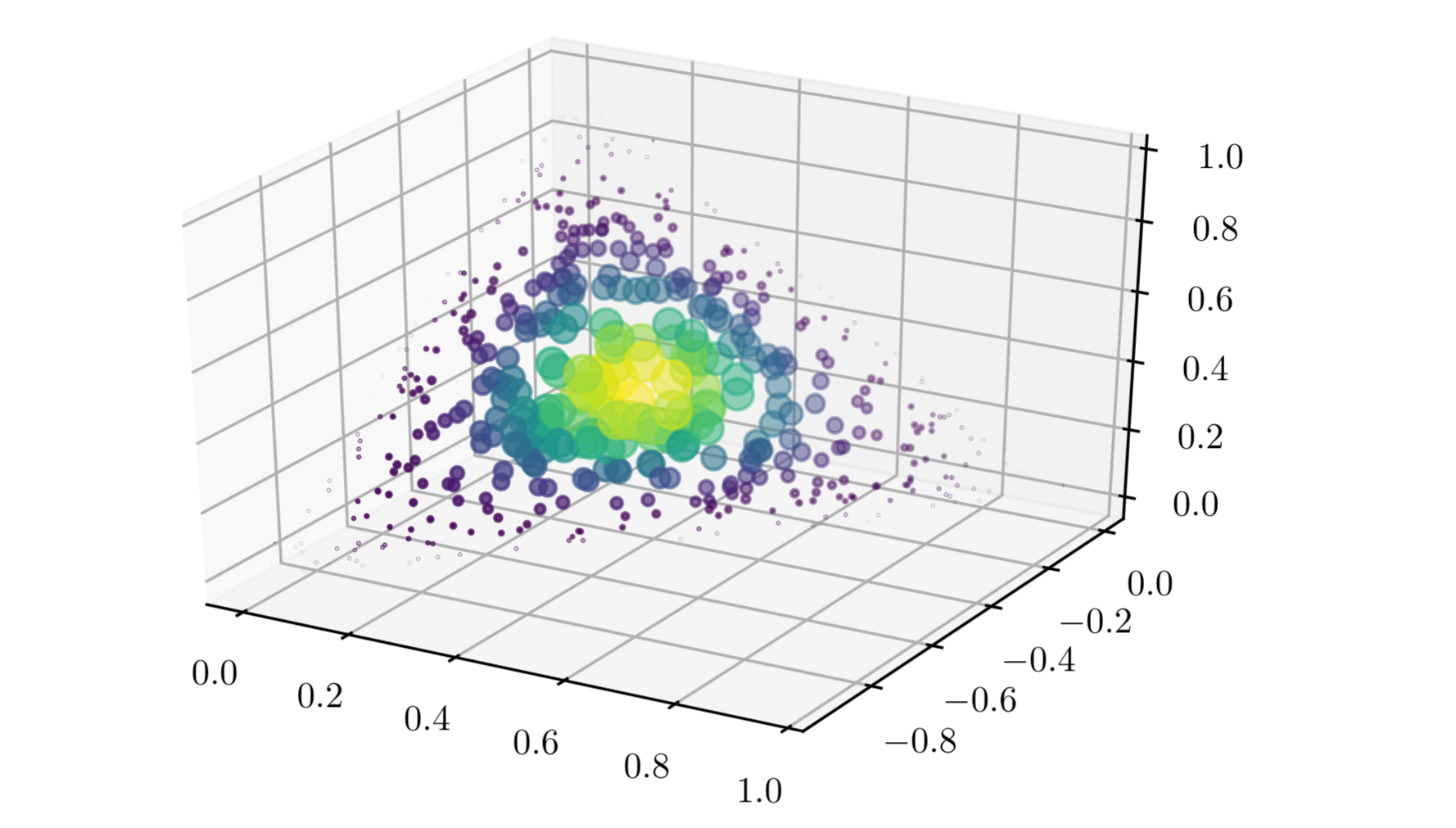}
  \caption{Dirichlet distributions of cardinality, $3$, visualised in 3D.
    For both (a) and (b), all $3$  $\bm{\alpha}$'s are equal.
    (\textbf{a}) Dirichlet with $\bm{\alpha} = \{0.5,0.5,0.5\}$.
    There is more mass on the corners of all $3$ axes than in the middle of the 
    hyperplane (sparse).
    (\textbf{b}) Dirichlet with $\bm{\alpha} = \{5,5,5\}$.
    There is less mass on the corners of all $3$ axes than in the middle of the 
    hyperplane. 
    \label{fig:dirsb}}
\end{figure}

Figures \ref{fig:dirsa} and \ref{fig:dirsb} illustrate Dirichlet distributions 
of cardinality $K = 3$ to illustrate the effect of $\bm{\alpha}$ on the 
distribution. Low values of $\alpha_k$ correspond to a low bias towards the 
corresponding $\theta_k$ parameter. These biases are also called pseudocounts.

\begin{figure}[H]
\vskip 0.2in
\begin{center}
\centerline{\includegraphics[width=0.6\columnwidth]{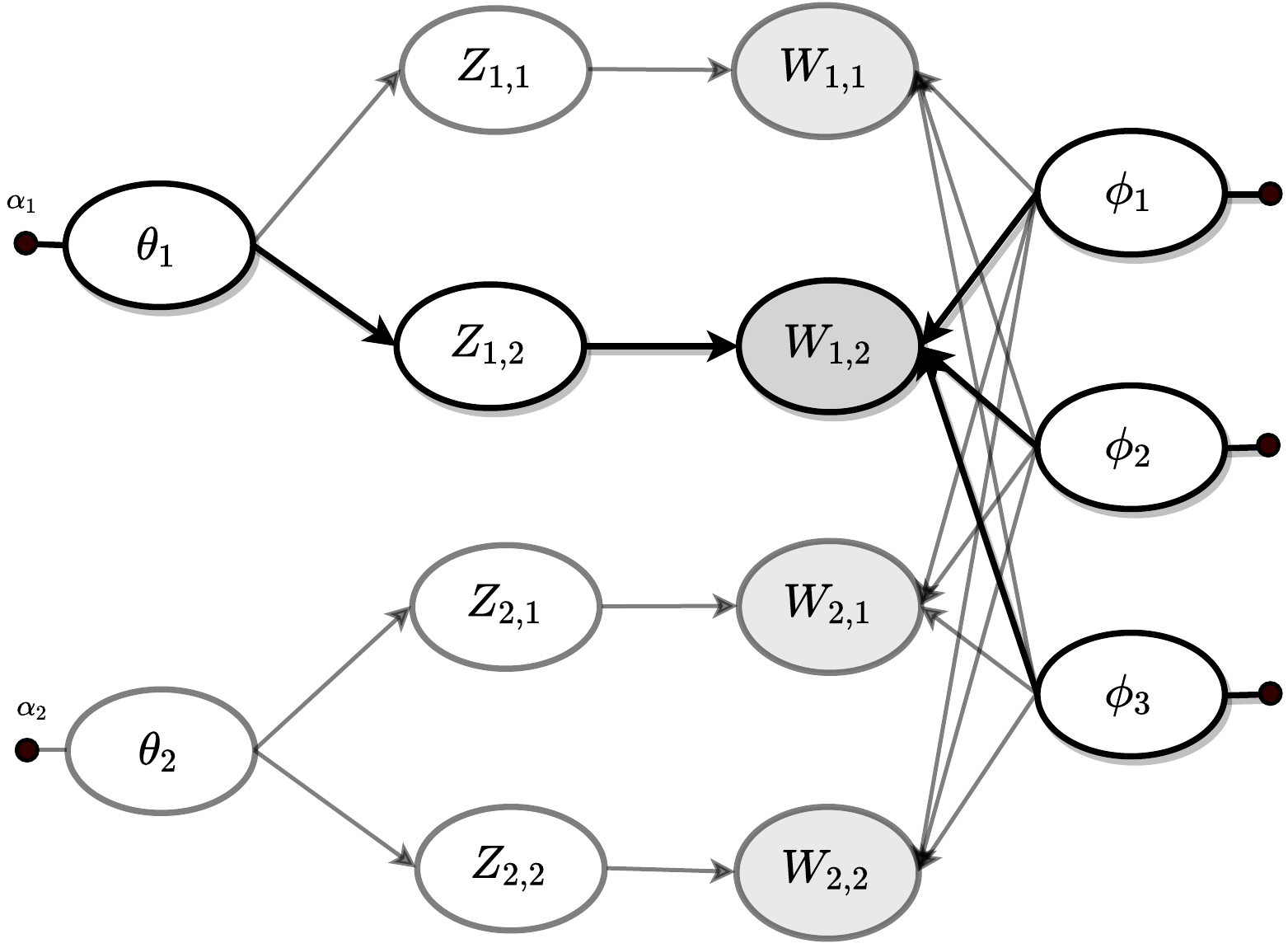}}
\caption{Unrolled BN representation of LDA for a two document corpus with two words per document. A single \textit{branch} is highlighted (where the document number is $m =1$ and the word number $n = 2$). For the meanings of the symbols, refer to Table \ref{table:symbols}. \label{fig:LDAbn_branch_hilightcg} }
\end{center}
\vskip -0.2in
\end{figure}

\begin{table}[H]
\small
 \caption{Symbols used for the LDA model shown in Figure~\ref{fig:LDAbn_branch_hilightcg}. \label{table:symbols}}
\begin{center}
\begin{tabular}{|c|  l|} 
 \hline
 Symbol & Description\\ [0.5ex] 
 \hline
 $M$ & Total  number of documents \cr $m$ & Current document \\ 
 \hline
 $N$ & Number of words in current document\cr $n$ & Current word (in document)\\
 \hline
 $K$ &  Total number of topics\cr $k$ & Current topic\\
 \hline
 $V$ & Total number words in the vocabulary \cr $v$ & Current word (in vocabulary)\cr $\mathsf{v}$ & Observed word (in vocabulary)\\
   \hline
  $\bm{\theta}_m$ & Topic-document Dirichlet for document $m$\\
  $Z_{m,n}$ & Topic-document categorical for word $n$ in document $m$\\
   $W_{m,n}$ & Word-topic conditional categorical for word $n$ in document $m$\\
     $\bm{\phi}_k$ & Word-topic Dirichlet for topic $k$ \\
  \hline
\end{tabular}
\end{center}
\end{table}

In LDA, the topic-word Dirichlet distributions can range 
from as low as $K = 3$ for a three topic model, up to very large values of $K$ 
for models containing hundreds of topics. The word-topic Dirichlet distributions 
typically have a much higher cardinality, typically in the thousands or hundreds 
of thousands since it corresponds to the vocabulary size.

\subsection{The exponential family (EF) of distributions}
\label{sec:expfamily}
The standard VMP algorithm is limited to distributions in the exponential family.
The exponential family is the only family of distributions with finite-sized 
sufficient statistics \cite{zamzami2020sparse,zamzami2020high,pan2020bayesian}.
Many useful distributions, including the Dirichlet and categorical distributions,
fall into this family, as do all the distributions involved in LDA.

The probability distribution of a random vector $\bm{x}$ with parameters
$\bm{\eta}$ can always be written in the following mathematically convenient 
form if it falls within the exponential family \cite{winn2004variational},
\begin{align}
  p(\bm{x};\bm{\eta})
   & = \frac{1}{Z(\bm{\eta})}h(\bm{x}) \exp
  \left\{
  \bm{\eta}^T\bm{T}(\bm{x})
  \right\},\label{eq:exp}
\end{align}
where $\bm{\eta}$ is called the natural
parameters and $\bm{T}(\bm{x})$ is the sufficient statistics vector,
called so since with a sufficiently large sample, the probability of $\bm{x}$ 
under $\bm{\eta}$ only depends on $\bm{x}$ through $\bm{T}(\bm{x})$.
The partition function, $Z(\bm{\eta})$, normalises
the distribution to unity volume i.e.
\begin{align}
  Z(\bm{\eta}) = \int h(\bm{x}) \exp
  \left\{
  \bm{\eta}^T\bm{T}(\bm{x})
  \right\}\text{d}\bm{x}. \label{eq:EFnorm}
\end{align}

We can also formulate Equation \ref{eq:exp} as,
\begin{align}
  p(\bm{x};\bm{\eta})
           & = h(\bm{x}) \exp
  \left\{
  \bm{\eta}^T\bm{T}(\bm{x}) - A(\bm{\eta})
  \right\}, &                  & \text{with } A(\bm{\eta}) \triangleq \log Z(\bm{\eta}).
\end{align}

$A(\bm{\eta})$ is known as the log-partition or cumulant
function since it can be used to find the cumulants of a distribution.
Below we show the first cumulant (the mean) of exponential family distributions 
which will be used later,
\begin{align}
  \nabla_{\bm{\eta}} A(\bm{\eta})
   & =  \nabla_{\bm{\eta}}\ln \left[\int h(\bm{x}) \exp
    \left\{
    \bm{\eta}^T\bm{T}(\bm{x})
  \right\}\text{d}\bm{x}\right] \nonumber                   \\
   & = \frac{1}{\int h(\bm{x}) \exp
    \left\{
    \bm{\eta}^T\bm{T}(\bm{x})
    \right\}\text{d}\bm{x}}\nabla_{\bm{\eta}} \left[\int h(\bm{x}) \exp
    \left\{
    \bm{\eta}^T\bm{T}(\bm{x})
  \right\}\text{d}\bm{x}\right]\nonumber                    \\
   & = \int \frac{1}{Z(\bm{\eta})}h(\bm{x}) \exp
  \left\{
  \bm{\eta}^T\bm{T}(\bm{x})
  \right\}\bm{T}(\bm{x})\text{d}\bm{x} \nonumber            \\
   & = \int p(\bm{x})\bm{T}(\bm{x})\text{d}\bm{x} \nonumber \\
   & = \left<\bm{T}(\bm{x})\right>_{p(\bm{x})}
  \label{eq:dlogpart}
\end{align}
This shows that we can find the first expected moment of a distribution in the 
exponential family by taking the derivative of its log-partition function.
This will always be the same as finding the expected value of the sufficient 
statistics vector.

\section{Variational message passing (VMP)}
\label{sec:vmp}
Variational Bayes (VB) is a framework for approximating the full posterior 
distribution over a model's parameters and latent variables in an iterative,
Expectation Maximization (EM)-like manner \cite{attias2000variational}, since the true distribution can often not be calculated efficiently for models of interest, such as LDA.

Variational message passing (VMP) is a way to derive the VB update equations 
for a given model \cite{winn2004variational, winn2005variational}. 
A formulation of optimization in terms of local computations is required to 
translate VB into its message passing variant, VMP.

\subsection{The generic VMP algorithm}
\label{sec:genvmp}
Conjugate-exponential models, of which LDA is an example, are models where conjugacy exists between all 
parent-child relationships and where all distributions are in the exponential 
family.
For these models we can perform variational inference by performing local 
computations and message passing.

\begin{figure}[H]
  \centering
  \includegraphics[width=100mm]{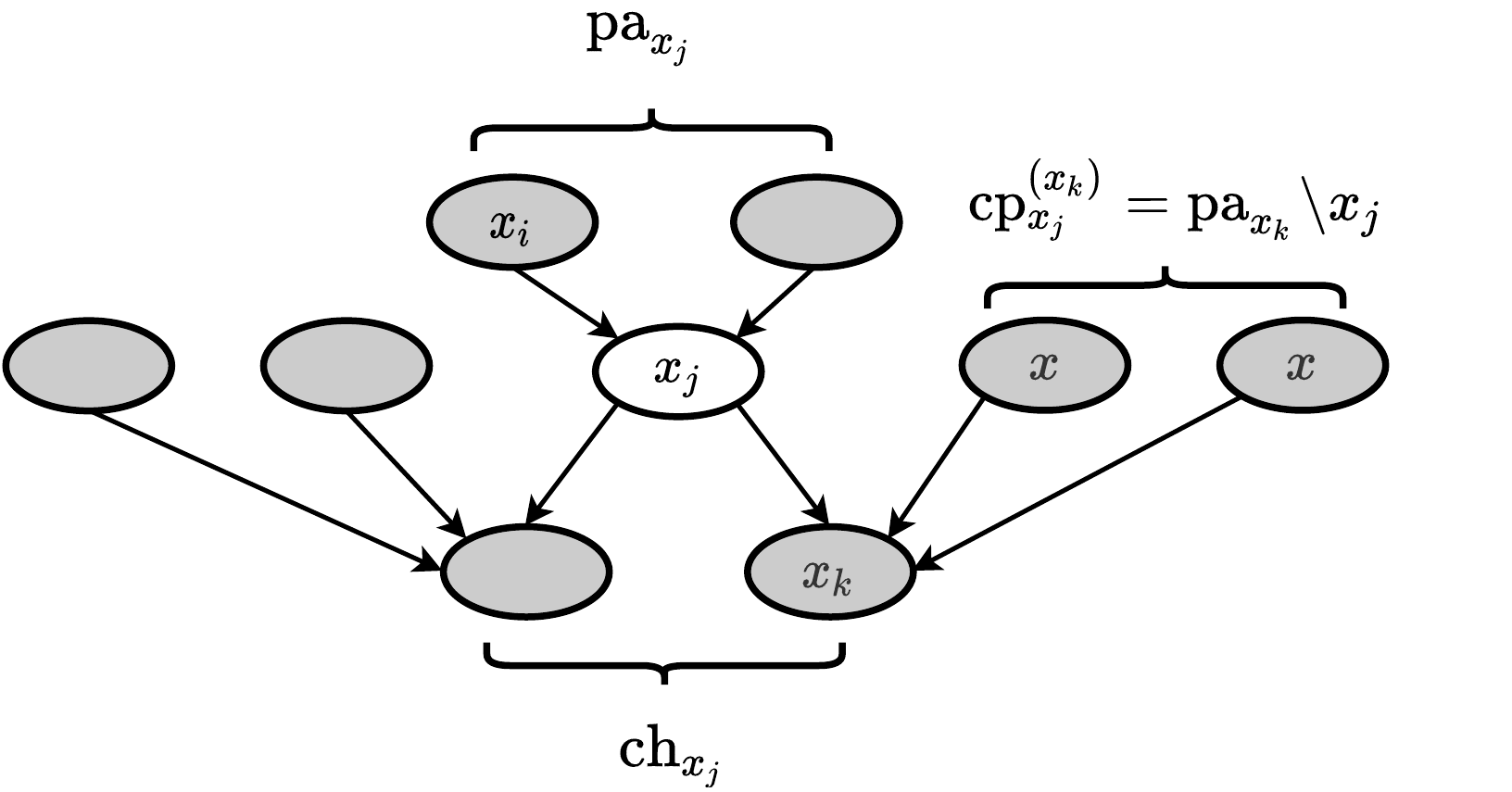}
  \caption{The Markov blanket of a node ($x_j$). 
   The nodes in the Markov blanket are the shaded nodes and are defined by the set 
   of parents (pa), children (ch) and co-parents (cp) of a node.
   The variational update equation for a node $x_j$ depends only on expectations 
   over variables appearing in the Markov blanket of that 
   node~\cite{winn2005variational}. 
   \label{fig:markov}}
\end{figure}

Based on the derivation in \cite{winn2004variational}, these local computations 
depend only on the variables within the Markov blanket of a node.
The nodes in the Markov blanket (shown in Figure \ref{fig:markov}) are nodes 
that are either parents, children, or co-parents of the node.
The co-parents of a node are the parents of its children and excludes the node 
itself.

For Bayesian networks, the joint distribution
can be expressed in terms of the conditional distributions at each node $x_i$,
\begin{align}
  p(\bm{x})= \prod_i p(x_i\mid \text{pa}_{x_i}), 
  \label{eq:bn01}
\end{align}
where $\text{pa}_{x_i}$ are the parents of node $x_i$ and $x_i$ the
variables associated with node $x_i$~\cite{koller2009probabilistic}. These variables can be either hidden, meaning we don't know, or or 
observed, meaning we know their values at inference.

\begin{figure}[H]
  \centering
  \includegraphics[width=20mm]{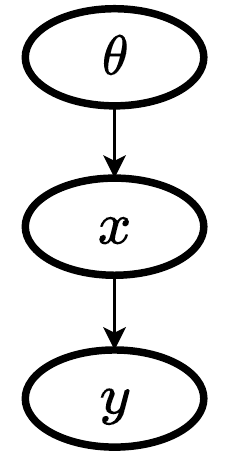}
  \caption{The child and parent nodes in the example we will use to present the 
   message passing equations.
   In the example, $\bm{x}$ is the parent node of $\bm{y}$, and $\bm{y}$ the child of $\bm{x}$; the node names correspond to the random vectors. 
   \label{fig:vmp}}
\end{figure}

In this section we review the general VMP message passing update equations and 
apply them to the example graphical model shown in Figure \ref{fig:vmp} to 
explain the broader principles involved in VMP.

The node $\bm{x}$, which represents a random vector is the parent of $\bm{y}$. In exponential family form this is written as,
\begin{align}
  \label{eq:exponentialfamily}
  p(\bm{x};\bm{\eta})
           & = \frac{1}{Z(\bm{\eta})}h(\bm{x}) \exp
  \left\{
  \bm{\eta}^T\bm{T}(\bm{x})
  \right\},                                                                  \\
           & = h(\bm{x}) \exp
  \left\{
  \bm{\eta}^T\bm{T}(\bm{x}) - A(\bm{\eta})
  \right\}, &                                        & \text{with } A(\bm{\eta}) \triangleq \ln Z(\bm{\eta}),
\end{align}
where $\bm{\eta}$ are the natural parameters, $\bm{T}(\bm{x})$ the sufficient 
statistics vector, and $A(\bm{\eta})$ the log-partition function.

If we limit ourselves to distributions in this family, the prior and posterior 
distributions have the same form \cite{winn2004variational}.
During inference, we therefore only need to update the values of the parameters 
and do not have to change the functional form \cite{winn2004variational}.

\subsubsection{Message to a child node}

Continuing with the description of the graphical model in Figure \ref{fig:vmp},
the parent to child node message for parent node  $\bm{x}$ and child node $\bm{y}$ is the expectation of the sufficient statistics vector 
\cite{winn2005variational},
\begin{equation}
  \bm{\mu}^{\text{p2c}}_{\bm{x}\rightarrow \bm{y}}  =\left<\bm{T}(\bm{x})\right>_{p(\bm{x})}.
  \label{eq:p2c}
\end{equation}

We can calculate this by using the derivative of the log-partition function as 
seen in Equation \ref{eq:dlogpart},
\begin{equation}
  \left<\bm{T}(\bm{x})\right>_{p(\bm{x})} = \nabla_{\bm{\eta}} A(\bm{\eta}).
  \label{eq:derivexpt}
\end{equation}

The parent to child message is therefore,
\begin{equation}
  \bm{\mu}^{\text{p2c}}_{\bm{x}\rightarrow\bm{y}} = \nabla_{\bm{\eta}} A(\bm{\eta}).
\end{equation}

This message which contains the expected values of the natural parameters of 
$\bm{x}$ now becomes the new natural parameters of $\bm{y}$. 
We can therefore write the child node's distribution as 
$p(\bm{y}\mid \left<\bm{T}(\bm{x})\right>)$ after receiving a message from node $x$.

\subsubsection{Message to a parent node}
Because we limit ourselves to conjugate-exponential models, the exponential form 
of the child distribution can always be re-arranged to match that of the parent 
distribution. 
This is due to the multi-linear properties of conjugate-exponential models 
\cite{winn2004variational}.

We define the re-arranged version of the sufficient statistics as 
$\bm{\varphi}(\bm{y})$.
This version is in the correct functional form to send a child to parent message  \cite{winn2004variational, winn2005variational}.
A child to parent message can therefore be written as,
\begin{equation}
  \label{eq:c2p}
  \bm{\mu}^{\text{c2p}}_{\bm{y}\rightarrow \bm{x}} =\left<\bm{\varphi}(\bm{y})\right>_{p(\bm{y})}.
\end{equation}

Note that if any node, ${a}$, is observed then the messages are as defined above,
but with $\left<\bm{\varphi}(\bm{a})\right>$ replaced by $\bm{\varphi}(\bm{a})$.
I.e. if we know the true values, we use them.

When a parent node has received all of its required messages, we can update its belief by finding its updated natural parameter vector $\bm{\eta}'$. In the general case for a graphical model containing a set of nodes $\bm{x} = \set{{x}_1,{x}_2,...,{x}_U}$, the update for parent ${x}_j$ becomes,
\begin{equation}\
  \label{eq:etaupdate}
  \bm{\eta}'_{{x_j}}=\left\{\bm{\mu}_{{x}_i\rightarrow {x}_j} \right\}_{{x}_i\in \text{pa}_{{x}_j}} + \sum_{s\in \text{ch}_{{x}_j}}\bm{\mu}_{{x}_s\rightarrow {x}_j}.
\end{equation}
Updating the parent node $\bm{x}$ from the graphical model in Figure~\ref{fig:vmp} will result in the update equation $\bm{\eta}'_{\bm{x}}=\bm{\mu}_{\bm{\theta}\rightarrow \bm{x}} + \bm{\mu}_{\bm{y}\rightarrow\bm{x}}$.

We now present the full VMP algorithm in~Algorithm~\ref{alg:VMP} as given by Winn~\cite{winn2004variational} but using our notation defined above.

\begin{algorithm}[H]
  \caption{Variational Message Passing (VMP) \cite{winn2004variational}}
  \label{alg:VMP}
  \begin{description}
    \item[Initialization:] ~\\Initialize each factor distribution $\bm{q}_j$ 
      by initializing the corresponding moment vector 
      $\left<\bm{T}_{j}(\bm{x}_j)\right>$ with random vector $\bm{x}_j$.
    \item[Iteration:]~\\[-2ex]
      \begin{enumerate}
        \item For each node $\bm{x}_j$ in turn:
              \begin{enumerate}
                \item Retrieve messages from all parent and child nodes, as 
                  defined in Equation \ref{eq:p2c} and Equation \ref{eq:c2p}.
                  This requires child nodes to retrieve messages from the 
                  co-parents of $\bm{x}_j$ (Figure \ref{fig:markov}).
                \item Compute updated natural parameter vector  
                  $\bm{\eta}'_j$ using Equation \ref{eq:etaupdate}.
                \item Compute updated moment vector 
                  $\left<\bm{T}_{j}(\bm{x}_j)\right>$ given the new setting of 
                  the parameter vector.
              \end{enumerate}
          $\text{ELBO}(\bm{q})$ (optional).
      \end{enumerate}
      \item[Termination:]~\\ If the increase in the bound is negligible or a 
        specified number of iterations has been reached, stop. 
        Otherwise, repeat from step 1.
  \end{description}
\end{algorithm}

\section{The VMP algorithm for LDA}
Here we describe the distribution at each node and also derive the child to 
parent and parent to child messages (where applicable) for the LDA graph.
We keep the messages dependent only on the current round of message passing 
which can be considered to be one epoch.
\label{sec:vpmlda}
\subsection{Topic-document Dirichlet nodes $\bm{\theta}_m$ }

In exponential family form we can write each topic-document Dirichlet as,
\begin{align}
  \ln \text{Dir}(\bm{\theta}_{{m}};\bm{\alpha}_{{m}})
  &=
  \left[
  \begin{array}{c}
    \alpha_{{m},1} - 1\\
    \alpha_{{m},2} - 1\\
    \vdots \\
    \alpha_{{m},K} - 1
  \end{array}
  \right]^T
  \left[
  \begin{array}{c}
    \ln \theta_{{m},1} \\
    \ln \theta_{{m},2} \\
    \vdots \\
    \ln \theta_{m,K}
  \end{array}
  \right]
 - \ln\frac{\Gamma(\sum_{k} \alpha_{{m},k})}{\prod_{k}\Gamma(\alpha_{{m},k})}. \label{eq:dirichletEFdoc}
\end{align}
For each $m$ we can identify the natural parameters as,
\begin{equation}
 \bm{\eta}_{\bm{\theta}_{{m}}} =  \left[
  \begin{array}{c}
    \alpha_{m,1}-1\\
    \alpha_{m,2}-1\\
    \vdots \\
    \alpha_{m,K}-1
  \end{array}
  \right],
\end{equation}
and the sufficient statistics as,
\begin{equation}
\bm{T}(\bm{\theta}_m) =
  \left[
  \begin{array}{c}
    \ln \theta_{m,1} \\
    \ln \theta_{m,2} \\
    \vdots \\
    \ln \theta_{m,K}
  \end{array}
  \right].\label{eq:dirsuff}
\end{equation}

\subsubsection{Message to a child node $Z_{m,n}$}
The parent to child node message (for parent node $\bm{\theta}_m$ and child node $ Z_{m,n}$) is the expectation of the sufficient statistics vector (Equation~\ref{eq:p2c}),
\begin{equation}
  \bm{\mu}^{\text{p2c}}_{\bm{\theta}_m\rightarrow  Z_{w,n}} =\left<\bm{T}(\bm{\theta}_m)\right>_{p(\bm{\theta}_m)}.
  \label{eq:parent2child}
\end{equation}
Using Equation~\ref{eq:derivexpt}, we can calculate this expectation using the derivative of the log-partition function. It is shown in~\cite[p128]{winn2004variational} to be,
\begin{align}
  \left<\ln(\theta_k)\right>_{p(\theta_k)} &= \psi(\alpha_k) - \psi(\sum_k\alpha_k), \label{eq:dir_E_suffstats}
\end{align}
where $\psi$ is the digamma function. The parent to child message from each topic-document Dirichlet node is therefore,
\begin{equation}
 \bm{\mu}^{\text{p2c}}_{\bm{\theta}_{m}\rightarrow  Z_{{m},n}} =  \left[ \begin{array}{c}
    \psi(\alpha_{{m},1}) - \psi(\sum_k\alpha_{{m},k})\\
    \vdots\\
    \psi(\alpha_{{m},K}) - \psi(\sum_k\alpha_{{m},k})\end{array} \right]. \label{eq:digamma1}
\end{equation}

We can now insert these expected sufficient statistics at the child node. The natural parameter vector is then,
\begin{equation}
\bm{\eta}'_{ Z_{{m},n}}
 = \left[ \begin{array}{c}
  \psi(\alpha_{{m},1}) - \psi(\sum_k\alpha_{{m},k}) \\
\vdots\\
 \psi(\alpha_{{m},K}) - \psi(\sum_k\alpha_{{m},k})\end{array} \right]
 = \left[ \begin{array}{c}
  \ln \theta^{'}_{{m},1}\\
\vdots\\
 \ln \theta^{'}_{{m}, K}\end{array} \right], \label{eq:catupdate1}
\end{equation}
with $\bm{\theta}'$ denoting the updated topic proportions, and $\bm{\eta}'_{ Z_{{m},n}}$ the updated natural parameter vector. Because these are not inherently normalised, normalisation is required to represent them as true probability distributions. Note also the conjugacy between the Dirichlet and categorical distributions: this allows us to simply update the natural parameters without changing the form of the distribution~\cite{masegosa2017scaling, winn2004variational}.

\subsection{Topic-document categorical nodes $Z_{m,n}$}
In exponential form we can represent the topic-document categorical distribution for a specific word in a specific document as,
\begin{align}
  \ln  \text{Cat}( Z_{{m},n}\mid \bm{\theta}_{{m}})  &=
  \left[\begin{array}{c}
    \ln \theta_{{m},1} \\
    \ln \theta_{{m},2} \\
    \vdots \\
    \ln \theta_{m,k}
  \end{array}
  \right]^T
  \left[\begin{array}{c}
    \llbracket Z_{{m},n} =1 \rrbracket \\ 
    \llbracket Z_{{m},n} =2 \rrbracket \\
    \vdots \\
    \llbracket Z_{{m},n} =K \rrbracket
  \end{array}
  \right]
  -\ln(\sum_k\theta_{m,k}),\label{eq:catEFa} \\
  \therefore A(\bm{\theta}_{{m}}) & =  \ln(\sum_k\theta_{m,k}), \label{eq:logpartc2}
\end{align}
by applying Equation~\ref{eq:exponentialfamily}.

For each branch (as defined in Figure~\ref{fig:LDAbn_branch_hilightcg}) we can identify the natural parameters as,
\begin{equation}
 \bm{\eta}_{Z_{{m},n}} =  \left[
 \begin{array}{c}
    \ln \theta_{{m},1} \\
    \ln \theta_{{m},2} \\
    \vdots \\
    \ln \theta_{m,k}
  \end{array}
  \right],
\end{equation}
and the sufficient statistics as,
\begin{equation}
\bm{T}(Z_{{m},n}) =
  \left[
  \begin{array}{c}
    \llbracket Z_{{m},n} =1 \rrbracket \\
    \llbracket Z_{{m},n} =2 \rrbracket \\
    \vdots \\
    \llbracket Z_{{m},n} =K \rrbracket
  \end{array}
  \right].\label{eq:dirsuffz}
\end{equation}

\subsubsection{Message to a parent node $\bm{\theta}_m$}
\label{sec:thetamsubs}
Before deriving the message from child node $Z_{{m},{n}}$ to parent node $\bm{\theta}_m$, some discussion regarding the effect of the incoming message from node $W_{{m},{n}}$ on node $Z_{{m},{n}}$ is in order. Because this message to the topic-document node  $Z_{{m},{n}}$ (from the respective word-topic node $W_{{m},{n}}$) is a child to parent message, this message is added to the natural parameter vector such that,
\begin{align}
    \ln\bm{\tilde{\theta}}^{'}_{{m}} &= \ln \bm\theta^{}_{{m}} + \ln \bm{p}_{{m},{n}}\notag\\
    &= \ln (\bm\theta^{}_{{m}}\bm{p}_{{m},{n}}), \label{eq:thetapxnonorm}
\end{align}
with $\tilde{.}$ indicating that the factor is unnormalised. Re-normalizing so that $\sum_k\theta^{'}_{{m},k} = 1$ gives,
\begin{align}
    \ln\bm{{\theta}}^{'}_{{m}}&= \ln (\frac{\bm\theta^{}_{{m}}\bm{p}_{{m},{n}}}{\sum_k\theta^{}_{{m},k}{p}_{{m},{n},k}}), \label{eq:thetapx}
\end{align}
where $\{p_{{m},{n},1}, ..., p_{{m},{n},K}\}$ are the topic probabilities for a specific word in a specific document as given by the message from node ${W}_{{m},{n}}$. Once each topic-document node $Z_{{m},{n}}$ has received its message from the corresponding word-topic node $W_{m,n}$, the natural parameter vector at node $Z_{{m},{n}}$ (from Equation~\ref{eq:catEFa}) will have been modified to be $\bm{\eta}'_{Z_{{m},n}} = \ln\bm{\theta}'_{{m}}$.

In the case where no message has been received from node ${W}_{{m},{n}}$, then  $\ln\bm{\theta}'_{{m}} =\ln\bm{\theta}^{}_{{m}}$. In LDA, however, we will only ever update the topic-document Dirichlet node $\bm{\theta}_m$ from the topic-document node $Z_{{m},{n}}$  after receiving a message from the word-topic side of the graph (except at initialisation).

The message from a topic-document node $Z_{{m},{n}}$ to a topic-document node $\bm{\theta}_m$ is also a child to parent message. To send a child to parent message we need to rearrange the exponential form of the child distribution to match the parent distribution (as presented in Section~\ref{sec:genvmp}). We can rearrange Equation~\ref{eq:catEFa} in terms of $\bm{\theta}'_{{m}}$ as follows,
\begin{align}
  \ln \text{Cat}({Z}_{{m},{n}}\mid \bm{\theta}_{{m}})  &=
 \left[\begin{array}{c}
    \llbracket Z_{{m},{n}} =1 \rrbracket \\ 
    \llbracket Z_{{m},{n}} =2 \rrbracket \\
    \vdots \\
    \llbracket Z_{{m},{n}} =K \rrbracket
  \end{array}
  \right]^T\left[ \begin{array}{c}
  \ln \theta^{'}_{{m},1}  \\
  \ln \theta^{'}_{{m},2}\\
\vdots\\
 \ln \theta^{'}_{{m}, K}\end{array} \right]
  -\ln(\sum_k\theta^{'}_{m,k}).\label{eq:catEF2}
\end{align}
The message towards node $\bm{\theta}_m$ is therefore,
\begin{align}
\bm{\mu}^{\text{c2p}}_{{Z}_{{m},{n}}\rightarrow \bm{\theta}_{{m}}} &=\left<\varphi({Z}_{{m},{n}})\right>_{p({Z}_{{m},{n}})}\notag\\
&=
 \left<\left[\begin{array}{c}
    \llbracket Z_{{m},{n}} =1 \rrbracket \\ 
    \llbracket Z_{{m},{n}} =2 \rrbracket \\
    \vdots \\
    \llbracket Z_{{m},{n}} =K \rrbracket
  \end{array}
  \right]\right>_{p({Z}_{{m},{n}})}
  =\left[ \begin{array}{c}
  \theta^{'}_{{m},1} \\
  \theta^{'}_{{m},2} \\
\vdots\\
 \theta^{'}_{{m}, K}.\end{array} \right]
 = \left[ \begin{array}{c}
  \frac{\theta^{}_{{m,1}}{p}_{{m},{n},1}}{\sum_k\theta^{}_{{m},k},{p}_{{m},{n},k}} \\
  \frac{\theta^{}_{{m,2}}{p}_{{m},{n},2}}{\sum_k\theta^{}_{{m},k}{p}_{{m},{n},k}}\\
\vdots\\
 \frac{\theta^{}_{{m,K}}{p}_{{m},{n},K}}{\sum_k\theta^{}_{{m},k}{p}_{{m},{n},k}}\end{array} \right].\label{eq:vecthetapx}
\end{align}
For each topic-document node $\bm\theta^{}_{m}$, for a single branch in the graph, and for a single topic ${k}$, we have the update,
\begin{equation}
    \alpha^{'}_{{m},{k}} =  \frac{\theta^{}_{{m,k}}{p}_{{m},{n},k}}{\sum_j\theta^{}_{{m},j}{p}_{{m},{n},j}} + \alpha^{\text{prior}}_{{m},{k}},\label{eq:approxmessagfin}
\end{equation}
with $\alpha^{\text{prior}}_{{m},{k}}$ representing the initial hyperparameter settings.
Over all words in the document we therefore have,
\begin{equation}
    \alpha^{'}_{{m},{k}} = \sum_n\frac{\theta^{}_{{m,k}}{p}_{{m},{n},k}}{\sum_j\theta^{}_{{m},j}{p}_{{m},{n},j}} + \alpha^{\text{prior}}_{{m},{k}},\label{eq:approxmessageAll1}
\end{equation}
which can also be written as,
\begin{equation}
    \alpha^{'}_{{m},{k}} = \sum_n\frac{\theta^{'}_{{m,k}}}{\sum_j\theta^{'}_{{m},j}} + \alpha^{\text{prior}}_{{m},{k}}.\label{eq:approxmessageAll1dash}
\end{equation}

\subsubsection{Message to a child node $W_{m,n}$}
The message from a topic-document node $Z_{m,n}$ to a word-topic node $W_{{m},n}$ is a parent to child message. Based on Equation~\ref{eq:p2c}, the message is,
\begin{equation}
\bm{\mu}^{\text{p2c}}_{{Z}_{{m},n} \rightarrow {W}_{{m},n}}
= \left<\bm{T}({Z}_{{m},n})\right>_{p({Z}_{{m},n})}.
\end{equation}
Equation~\ref{eq:exponentialfamily} defines the log-partition function that can be used to calculate this moment using Equation~\ref{eq:derivexpt}. To do this we need to re-parameterise the natural parameter vector from Equation~\ref{eq:catEFa}. This is shown below for a single word in a single topic $k$,
\begin{align*}
  \eta_k &\equiv \ln\theta'_k,\\
  \therefore \theta^{'}_k &= e^{\eta_k},\\
  \therefore A(\bm{\eta}) &= \ln(\sum_k e^{\eta_k}).
\end{align*}
From this we can calculate the expected sufficient statistics using
Equation~\ref{eq:dlogpart} for a specific topic ${k}$:
\begin{align}
  < \llbracket Z_{{m},n} =k \rrbracket >_{p({Z}_{{m},n})}
    &= \frac{\delta}{\delta\eta_{k}}A(\bm{\eta}) \nonumber \\
    &= \frac{e^{\eta_{k}}}{\sum_je^{\eta_{j}}} \nonumber \\
    &= \frac{\theta^{'}_{{m},{k}}}{\sum_j\theta^{'}_{{m},{j}}} \nonumber \\
    &= \theta^{*}_{{m},{{k}}},&& \text{where the $\theta^{*}$'s are normalised}. \label{eq:catSS}
\end{align}
Each $\theta^{*}_{{m},{k}}$ is a normalised topic proportion for a single topic. We can write the full parent to child message for a word within a topic as,
\begin{equation}
 \bm{\mu}^{p2c}_{{Z}_{m,n}\rightarrow {W}_{{m},n}} =  \left[ \begin{array}{c}
    \theta^{*}_{{m},1}\\
    \vdots\\
    \theta^{*}_{{m},K}\end{array} \right], \label{eq:digamma1a}
\end{equation}
with,
\begin{equation}
\theta^{*}_{{m},k} =\frac{\theta^{'}_{{m},{k}}}{\sum_j\theta^{'}_{{m},{j}}}.
\end{equation}
These updated topic proportions are then used at the word-topic node ${W}_{{m},n}$.

\subsection{Word-topic conditional categorical nodes $W_{m,n}$}

Initially, the $n$th word of a document is described by $K$ word-topic distributions. We call this a conditional categorical distribution; for a topic ${k}$ this reduces to a single categorical distribution. For all $K$ we can write,
\begin{align}
  \ln  \text{Cat}({W}_{{m},n}\mid {Z}_{{m},n},\bm{\Phi})  &=
  \left[
  \begin{array}{c}
    \sum_{k} \llbracket  Z_{{m},n} = {k}\rrbracket \ln \phi_{k,1} \\
    \sum_{k}\llbracket  Z_{{m},n} = {k}\rrbracket \ln \phi_{k,2}  \\
    \vdots \\
    \sum_{k}\llbracket  Z_{{m},n} = {k}\rrbracket  \ln \phi_{k,V}
  \end{array}
  \right]^T
  \left[
  \begin{array}{c}
    \llbracket W_{{m},n}=1\rrbracket \\ 
    \llbracket  W_{{m},n}=2\rrbracket \\
    \vdots \\
    \llbracket  W_{{m},n}=V\rrbracket
  \end{array}
  \right] \label{eq:catEF5} \\ \nonumber
  &-\sum_{k}\llbracket  Z_{{m},n} = {k}\rrbracket \ln(\sum_v \phi_{k,v}),\\ \nonumber
\end{align}
where the vocabulary over all words ranges from $1$ to $V$.

\subsubsection{Message to a categorical parent node $Z_{m,n}$}
Each word-topic node ${W}_{{m},{n}}$ is a child of a topic-document node ${Z}_{m,n}$; we therefore need to send child to parent messages between each pair of nodes. We can rewrite Equation~\ref{eq:catEF5} in terms of ${Z}_{m,n}$ to give,
  \begin{align}
   \ln  \text{Cat}({W}_{{m},{n}} \mid  {Z}_{{m},{n}},\bm{\Phi}) &=
    \left[
      \begin{array}{c}
        \sum_v \llbracket W_{m,n}=v\rrbracket\ln\phi_{1,v}\\
        \vdots \\
        \sum_v \llbracket W_{m,n}=v\rrbracket\ln\phi_{k,v}\\
        \vdots\\
        \sum_v \llbracket W_{m,n}=v\rrbracket\ln\phi_{K,v}
      \end{array}
      \right]^T
    \left[
      \begin{array}{c}
        \llbracket {Z}_{m,n}=1\rrbracket \\
        \vdots \\
        \llbracket {Z}_{m,n}=k\rrbracket \\
        \vdots \\
        \llbracket {Z}_{m,n}=K\rrbracket
      \end{array}
      \right]. \label{eq:ccatEFconjz3}
  \end{align}
After observing the word $\mathsf{v}$,  Equation~\ref{eq:ccatEFconjz3} reduces to a categorical form (this is always the case in standard LDA). Using Equation~\ref{eq:catSS}, we can then write the word-topic child to parent message as,
\begin{align}
\label{eq:parenttochildcat1_dup1}
  \bm{\widetilde{\mu}}^{\text{c2p}}_{{W}_{m,n} \rightarrow {Z}_{m,n}}
  &= \left[
      \begin{array}{c}
        \ln{\phi}_{1,\mathsf{v}}\\
        \vdots \\
        \ln{\phi}_{k,\mathsf{v}}\\
        \vdots\\
        \ln{\phi}_{K,\mathsf{v}}
      \end{array}
      \right],
\end{align}
where $\mathsf{v}$ is the observed word, and the message is unnormalised. This is because we have taken a slice through the word-topic distributions for a specific word, which means that the result is not a true distribution. We normalise the message to obtain the topic proportions (for each word in each document), which gives,
\begin{align}
\label{eq:parenttochildcat1_dup1norm}
  \bm{{\mu}}^{\text{c2p}}_{{W}_{m,n} \rightarrow {Z}_{m,n}}
  &= \left[
      \begin{array}{c}
        \ln\phi^{*}_{1,\mathsf{v}}\\
        \vdots \\
        \ln\phi^{*}_{k,\mathsf{v}}\\
        \vdots\\
        \ln\phi^{*}_{K,\mathsf{v}}
      \end{array}
      \right],
\end{align}
with,
\begin{equation}
\phi^{*}_{k,\mathsf{v}} = \frac{\phi_{k,\mathsf{v}}}{\sum_j\phi_{k,\mathsf{v}} }.
\end{equation}

To determine the updated document topic proportions we update the natural parameter vector by adding these topic weightings to the current document topic proportions,
\begin{equation}
\bm{\eta}'_{{Z}_{m,n}}
 = \left[ \begin{array}{c}
  \ln \theta^{}_{{m},1} + \ln\phi^{*}_{1,\mathsf{v}} \\
\vdots\\
 \ln \theta^{}_{{m}, K} + \ln\phi^{*}_{K,\mathsf{v}}\end{array} \right].\label{eq:catupdate2}
\end{equation}


\subsubsection{Message to a Dirichlet parent node $\bm{\phi}_k$}
To send child to parent messages from the a word-topic node ${W}_{{m},n}$ to each word-topic node $\bm{\phi}_k$, re-parameterisation is required.


After parameterisation in terms of $\phi_{k}$ we have,
  \begin{align}
   \label{eq:ccatEFphi}
    \ln  \text{Cat}({W}_{{m},n} \mid  {Z}_{{m},n},\bm{\phi}_{{k}}) &=
     \left[
      \begin{array}{c}
        \llbracket {Z}_{{m},n} = {k} \rrbracket  \llbracket  W_{{m},n}=1\rrbracket \\
        \vdots \\
        \llbracket {Z}_{{m},n} = {k} \rrbracket  \llbracket  W_{{m},n}=v\rrbracket \\
        \vdots \\
        \llbracket {Z}_{{m},n} = {k} \rrbracket  \llbracket  W_{{m},n}=V\rrbracket
      \end{array}
      \right]^T
    \left[
      \begin{array}{c}
        \ln\phi_{k,1}\\ 
        \vdots \\
        \ln\phi_{k,v}\\
        \vdots \\
        \ln\phi_{k,V}
      \end{array}
      \right] \\
      &+ \text{ terms involving } \ln\bm{\phi}_{{Z}_{{m},n} \ne {k}}, \nonumber
  \end{align}
where $\phi_{k,\mathsf{v}}$ are the topic proportions for word $\mathsf{v}$ in topic ${k}$.
The messages from one of these categorical beliefs to $\bm{\phi}_k$ can then be written as,
\begin{align}
  \bm{\mu}^{\text{c2p}}_{W_{{m},{n}}\rightarrow \bm{\phi}_{k}}
  &= \left<\varphi({W_{{m},{n}}, Z_{{m},n}=k})\right>_{p(W_{{m},{n}}, Z_{m,n}=k)}\nonumber\\
  &= \left<\llbracket Z_{{m},n}=k\rrbracket \llbracket{W}_{{m},{n}}=v\rrbracket \right>_{p({W}_{{m},{n}}, Z_{{m},n}=k)}\nonumber\\
  &= \left<\llbracket Z_{{m},n}=k\rrbracket\right>_{p({Z_{{m},n}=k)}}, \text{ because $W_{{m},{n}}$ is observed}.\nonumber\\
  \label{eq:wtophi3}
\end{align}
Note that because ${W}_{m,n}$ is observed, the values in the vector for all entries except for where ${W}_{{m},{n}}={v}$, are zero. To update $\bm{\phi}_{{k}}$, we simply add all the incoming message to the respective $\bm{\beta}_k$ values. For a specific word in the vocabulary ${v}$ this would be,
\begin{align}
\beta^{'}_{k,{v}} &= \sum_m\sum_n \theta^{*}_{{m},k} + \beta^{\text{prior}}_{k,{v}}\label{eq:betaupdate},
\end{align}
with $\theta^{*}_{{m},k}$ denoting the normalised probability of topic $k$ for document $m$.
We can see that the scaled topic proportions for word $\mathsf{v}$ are  simply added to the respective word's word-topic Dirichlet's parameters. 

\subsection{Word-topic Dirichlet nodes $\bm{\phi}_k$}
The word-topic distribution factors are of Dirichlet form. For the entire graph, we have: $\bm{\Phi} = \{\bm{\phi}_1, ..., \bm{\phi}_{K} \}$. For each topic $k$ we write,
\begin{align}
  \ln\text{Dir}(\bm{\phi}_{{k}};\bm{\beta}_{{k}})
  &=
  \left[
  \begin{array}{c}
    \beta_{{k},1} - 1\\
    \beta_{{k},2} - 1\\
    \vdots \\
    \beta_{{k},V} - 1
  \end{array}
  \right]^T
  \left[
  \begin{array}{c}
    \ln \phi_{{k},1} \\
    \ln \phi_{{k},2} \\
    \vdots \\
    \ln \phi_{{k},V}
  \end{array}
  \right]
 - \ln\frac{\Gamma(\sum_{v} \beta_{k,v})}{\prod_{v}\Gamma(\beta_{{k},v})}. \label{eq:dirichletEFword}
\end{align}

\subsubsection{Message to a child node $W_{m,n}$}
These messages are very similar to the ones on the topic-document side of the graph since they are also parent to child messages with each parent having a Dirichlet form.

For each topic ${k}$ the messages sent to all word-topic nodes $W_{{m},{n}}$ (one for each word in each topic) will be identical,
\begin{equation}
  \bm{\mu}^{\text{p2c}}_{\bm{\phi}_{{k}} \rightarrow W_{m,n}}
=  \left[ \begin{array}{c}
    \psi(\beta_{{k},1}) - \psi(\sum_v\beta_{{k},v})\\
    \vdots\\
    \psi(\beta_{{k},V}) - \psi(\sum_v\beta_{{k},v})\end{array} \right].
\end{equation}
The additional complexity comes from the fact that the child nodes $W_{{m},{n}}$ need to assimilate messages from $K$ Dirichlet distributions and not only from one, as in the topic-document side of the graph.

We now perform a similar update to the update seen in Equation~\ref{eq:catupdate1}, except that we have $K$ messages added instead of only one. For each ${k}$ we have,
\begin{equation}
\bm{\eta}'_{\bm{\phi}_{{k}}}
 = \left[ \begin{array}{c}
  \psi(\beta_{k,1}) - \psi(\sum_v\beta_{k,v})\\
\vdots\\
 \psi(\beta_{{k},V}) - \psi(\sum_v\beta_{{k},v})\end{array} \right]
 = \left[
      \begin{array}{c}
        \ln\phi^{'}_{k,1}\\ 
        \vdots \\
        \ln\phi^{'}_{k,V}
      \end{array}
      \right],\label{eq:ccatupdate}
\end{equation}
where $\phi^{'}$ denotes the updated values.

We have now presented the VMP message updates for each node in the LDA graphical model.

Using these messages, VMP for LDA can be implemented using a range of message passing schedules. In the next section we provide one such message passing schedule.

\section{Message passing schedule}

Because LDA has a simple, known structure per document, it is sensible to 
construct a fixed message passing schedule. This is not always the case for 
graphical models, for example in some cases we chose rather to base the schedule 
on message priority using divergence measures to prioritise messages according 
to their expected impact
\cite{Brink2016UsingPG,Louw2018APG,streicher2021strengthening,taylor2021albu}.

In Algorithm \ref{alg:albu}, we presented our proposed VMP message passing 
schedule for LDA. 
It is based on the message passing schedule of the approximate loopy belief 
update (ALBU) VMP implementation in \cite{taylor2021albu} that uses a form of 
belief propagation \cite{lauritzen1988local, koller2009probabilistic}[p364-366].
There are, of course, are many other variants that one could use.

\begin{algorithm}[tbh]
    \caption{Message passing schedule for LDA}  \label{alg:albu}
    \begin{description}
    \item[For each epoch:] ~\\[-5mm]
    \begin{itemize}
    \item[] \begin{description}
    \item[For each document $\bm{m}$:] ~\\[-4mm]
    \begin{itemize}
        \item[] \begin{description}
        \item[For each word $\bm{n}$ in document $\bm{m}$:] ~\\[-4mm]
    \begin{itemize}
    \item[--] send messages from each node $\bm{\phi}_k$ to node $W_{m,n}$
    \item[--] observe word $W_{m,n} = \mathsf{v}$
    \item[--] send message from node $W_{m,n}$ to node ${Z_{m,n}}$
    \item[--] send message from node  $Z_{m,n}$ to $\bm{\theta}_m$
     \end{itemize}
    \item[For each word $\bm{n}$ in document $\bm{m}$:] ~\\[-4mm]
    \begin{itemize}
    \item[--] send message from node $\bm{\theta}_m$ to node $Z_{m,n}$
    \item[--] send message from node $Z_{m,n}$ to node $W_{m,n}$
    \end{itemize}
 \end{description}
  \end{itemize}
  \item[For each word $n$ in each document $m$:] ~\\[-4mm]
 \begin{itemize}
    \item[--] send messages from node ${W_{m,n}}$ to each $\bm{\phi}_k$
    \end{itemize}
\end{description}
\end{itemize}
\end{description}
\end{algorithm}

Based on this schedule, as well as the message passing equations provided in \ref{sec:vmp}, VMP can be implemented for LDA.

\section{Conclusion and future work}
VMP, an elegant and tenable solution to inference problems, has not been 
presented in detail for the standard, smoothed LDA graphical model, 
which is surprising in view of its speed and ease of use.

In this article, we provided an introduction to variational message passing (VMP), the message passing equivalent of VB. We present the generic VMP algorithm and then applied VMP to the LDA graphical model. 
Finally we proposed a message passing schedule for VMP for LDA. 
For future work, we recommend that VMP and VB be compared in terms of execution time for LDA, and that alternative message passing schedules be investigated to improve execution time and convergence rate. We also recommend that the VMP equations for other, similar graphical models be derived and published in a similar manner.


\bibliographystyle{unsrt}

\end{document}